\documentclass{article}
\usepackage{spconf,amsmath,graphicx,amsfonts}

\usepackage[pagebackref=true,breaklinks=true,letterpaper=true,colorlinks,bookmarks=false]{hyperref}

\title{Data Driven Coded Aperture Design for Depth Recovery}
\name{Prasan A Shedligeri, Sreyas Mohan, Kaushik Mitra}
\address{Department of Electrical Engineering\\ Indian Institute of Technology Madras, Chennai, India}
\begin{document}
%
\maketitle
\begin{abstract}
Inserting a patterned occluder at the aperture of a camera lens has been shown to improve the recovery of depth map and all-focus image compared to a fully open aperture. However, design of the aperture pattern plays a very critical role. 
Previous approaches for designing aperture codes make simple assumptions on image distributions to obtain metrics for evaluating aperture codes. However, real images may not follow those assumptions and hence the designed code may not be optimal for them. To address this drawback we propose a data driven approach for learning the optimal aperture pattern to recover depth map from a single coded image. We propose a two stage architecture where, in the first stage we simulate coded aperture images from a training dataset of all-focus images and depth maps and in the second stage we recover the depth map using a deep neural network. We demonstrate that our learned aperture code performs better than previously designed codes even on code design metrics proposed by previous approaches.

\end{abstract}
\section{Introduction}
\label{sec:intro}
In coded aperture photography \cite{levin2007image, veeraraghavan2007dappled} a patterned occluder is inserted at the aperture of a conventional camera. This gives more accurate scene depth map and all-focus image as compared to that from a conventional circular aperture \cite{levin2007image, veeraraghavan2007dappled, zhou2009coded}. However the design of the aperture pattern is very critical to its performance. For example, a broadband code can recover the all-focus image but then depth recovery is not optimal \cite{zhou2009coded}. On the other hand, a pattern with zero-crossings in its Fourier domain is better for depth recovery, but not for recovering the all-focus image. Here, our goal is to find the optimal pattern for depth recovery from a single coded image. 

Searching for the best pattern is a non-trivial and challenging task. For example, if we choose to use a $N \times N$ binary pattern, then there are $2^{N^2}$ choices, which makes the problem combinatorial. Earlier attempts at searching for code assumed some statistical priors on the images like Gaussian prior on image gradients \cite{levin2007image} or $1/f$ law based image prior \cite{zhou2009coded} and went about designing the code based on these simplifying assumptions. While these methods were able to show improvement over random aperture codes, these codes suffered with several impairments, one of the notable ones being the inability to work on non-textured images. With the abundance of data and cheaper computing power, deep learning based methods are becoming increasingly popular in the field of computational photography. A very successful recent example is the design of sensor multiplexing pattern for image demosaicing using a deep learning framework \cite{chakrabarti2016learning}. While the sensor multiplexing codes till then relied on expert intuitions \cite{chakrabarti2014rethinking} or some image specific model, this strategy of code design enabled a much broader data driven approach, co-designing both the code and the recovery method simultaneously.

In this paper, we propose a data-driven code design technique for depth map recovery from a single coded aperture image. The novelty of our approach is primarily in the fact that we use a data-driven coded aperture design instead of assuming specific image models. We use a 2 stage architecture where the first stage simulates the coded aperture image and the second stage predicts the depth given the simulated image using a deep neural network. Thus unlike previous approaches, we let the network generalize and learn what is important for recovering depth from a coded image. Interestingly, the data-driven approach automatically optimizes the criteria considered in \cite{levin2007image}. Our learned code has high Kullback-Leibler (KL) divergence between distribution of images blurred with different scaled versions of the aperture code, which makes depth prediction more accurate \cite{levin2007image}. Moreover, we are able to recover depth map from non-textured images using this approach when all the previously considered methods fail in this task. Also, depth estimation techniques used in \cite{levin2007image,zhou2009coded} make use of a series of convolution and deconvolution operations which make them slow and computationally complex. On the other hand we use Convolutional Neural Network (CNN) for predicting depth which is faster and less computationally complex.

\begin{figure*}[t]
  \centering
  \includegraphics[height=0.3\textwidth]{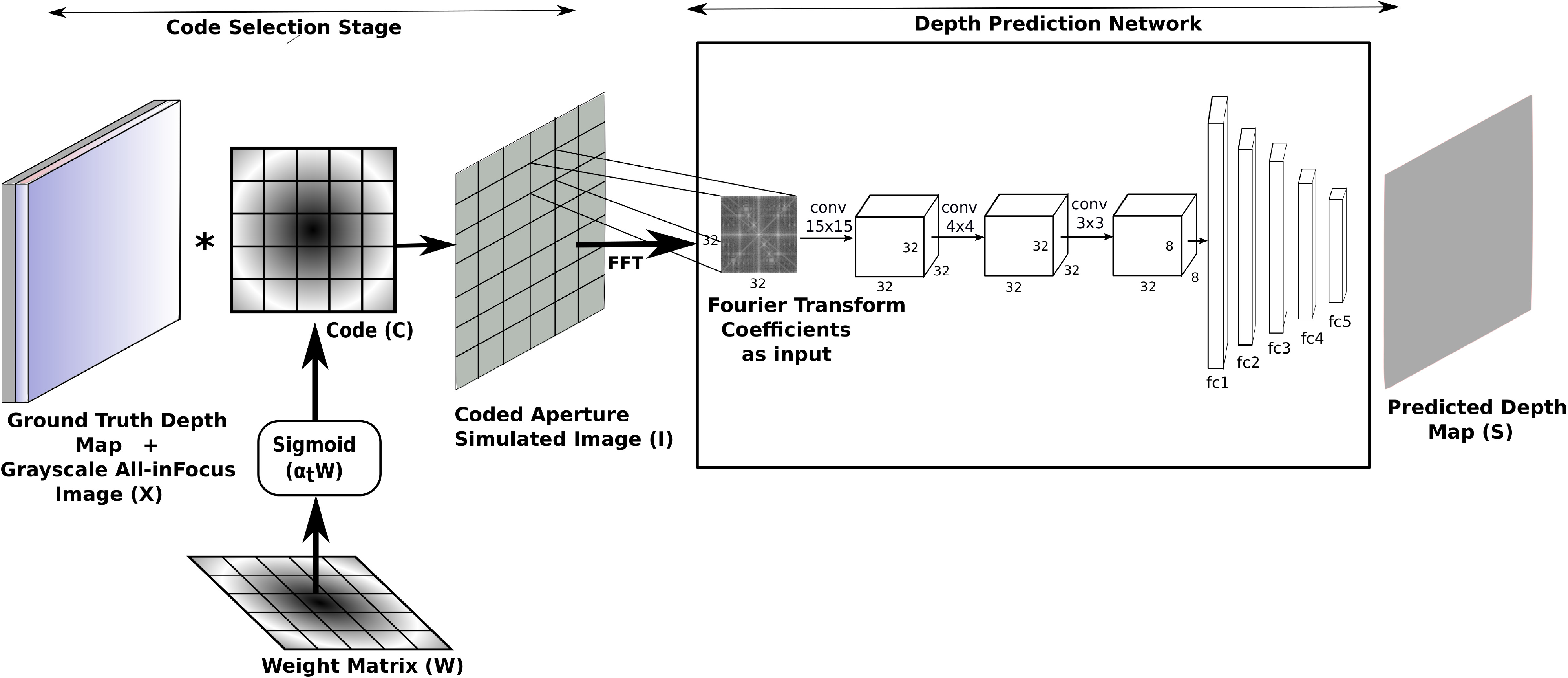}
  \caption{The framework used for learning the optimal code. Sigmoid is applied over the learnable weight matrix to obtain a code, which is convolved according to equation \eqref{eqn:simulatedImage} to obtain the coded aperture image. The coded image is then passed through the depth estimation network to obtain the depth map. The depth map prediction error is then back-propagated to update the network weights}
  \label{fig:net_architecture}
\end{figure*}

\section{Prior Work}
\label{sec:priorwork}

Depth estimation techniques can be broadly classified into active and passive methods. Active methods typically involve additional lighting sources and produces high quality estimates of depth map \cite{salvi2010state, axelsson1999processing, schwarte1997new}. Unlike active methods, passive methods capture the 3-dimensional information without any intervention and recovers the depth information by post-processing, usually by the analysis of changes in viewpoints or focus. One of the common passive approaches for recovering depth information is to compute depth from defocus. For any given camera setting, we will have only one plane in focus (focal plane). Points which are farther away from the focal plane will appear defocused and the amount of defocus will depend on the distance of the object from the focal plane. Therefore, the depth of a point can be found out by estimating its blur size in the image \cite{pentland1987new, subbarao1988depth, rajagopalan1997optimal, favaro2005geometric}.


Depth estimation techniques can also be classified as multi-image and single image based techniques. Multiple image capture like stereo pair \cite{scharstein2001taxonomy} or light field capture \cite{adelson1992single,ng2005light,levoy2006light} which involves capturing images from multiple view points. Multiple view point capture techniques rely on highly textured regions for a reliable depth estimate unlike the conventional real photographs which may lack textured regions. Single image depth estimation \cite{eigen2014depth,eigen2015predicting,saxena2005learning,garg2016unsupervised,liu2015deep} has been recently of interest with the advent of deep learning techniques. But these techniques usually do not consider defocus cue, which is one of the major depth cues.

Learning the optimal code for coded aperture imaging is a very critical and challenging task. In \cite{levin2007image}, the authors assume a Gaussian prior on the image gradients and formulate a cost function that evaluates the given pattern for depth estimation. The authors also point out that the zeros of the filter in Fourier domain should be non-overlapping across different scales. In \cite{zhou2009coded}, the authors try to learn two codes instead of one to efficiently extract depth information and all-focus image together. They assume an image prior based on $1/f$ law. In \cite{chakrabarti2016learning}, the author proposed a data driven approach to learn the optimal sensor multiplexing pattern for image demosaicing. We use a similar data driven approach for learning the optimal aperture code without making any assumptions on the image prior. 


\section{Framework for Learning Optimal Code}
\label{sec:pagestyle}

Inspired by the approach followed in \cite{chakrabarti2016learning}, our goal is to find the optimal aperture code as well as to learn a network that can predict the scene depth (in terms of blur kernel size) from the coded image. Note that the blur kernel structure is defined by the aperture code but the kernel size depends on the scene depth. Our architecture can be thought of having two stages. The first stage simulates the coded aperture image given an input all-focus image and the corresponding depth map. The second stage is a neural network that predicts the blur kernel size from the coded image; we call this network as the depth prediction network.


In order to simulate the defocus in the coded aperture image we need to know the size of the blur kernel with which to convolve the input image. The blur kernel size is constant only over a region of constant depth. Hence, we extract $32 \times32$ patches assuming that the depth stays constant in that patch. By knowing the depth at that patch we can calculate the size $s$ of the blur kernel. We then convolve the extracted image patch with a  blur kernel of size $s\times s$. Since, blur kernel is centered on a pixel the kernel size can take values of $1,3,5,\cdots k$ with k being the maximum blur size. Iterating this over all the patches and stitching the patches will give coded image as the output.

We denote the grayscale all-focus input images by $X$ and the output blur-size map by $S$, which is related to scene depth map. 
We denote the simulated coded image using $I$ and the aperture code with $C$, where, ${C} \in \{0, 1\}^{(N \times N)}$ is a $N\times N$ binary code. In our experiments we used $N=11$, i.e., $C$ is a binary $11 \times 11$ code.

Let $I^{(p)}$ and $X^{(p)}$ denote the $p^{th}$ patch in the coded image  and the input all focus image respectively. Then,
\begin{equation}
\label{eqn:simulatedImage}
    I^{(p)} = C^{(s)} \ast X^{(p)}
\end{equation}
where $C^{(s)}$ is the blur kernel scaled to appropriate size $s$.

The second stage of the network takes as input the Fourier transform of the patch of size $32\times 32$ extracted from the coded image and predicts the blur kernel size $s$. Since the kernel size values are discrete and finite, we formulate the scale prediction task as a classification problem with cross-entropy loss. The error in classifying the input patch is back-propagated to obtain the optimal pattern $C$.

\subsection{Code Prediction Stage}
\label{sec:codeprediction}


The goal here is to find the optimal aperture code $C$. Since, $I$ is simulated using the code $C$, any error at the kernel-size prediction can be attributed to the code $C$ as well as the weights of the depth prediction network. Hence, the gradients due to the cross-entropy error at the output will try to update the code such that the error is minimized. 

Instead of directly finding the optimal binary code $C$, we first optimize over real valued code $W \in \mathbb{R}^{N \times N}$ and then binarize it using sigmoid function. We define $ C = sigmoid[\alpha_t W] $, where $\alpha_t$ is a parameter which is linearly increased with number of iterations, $t$. With increasing value of $\alpha_t$ the sigmoid function becomes more steeper hence forcing the values of $C$ to saturate towards either 0 or 1. 
Even after applying the sigmoid function, in our experiments we found that $C$ is not exactly binary. Thus, to obtain a binary code, which is easier to realize physically, we threshold the obtained $C$. 

\subsection{Depth Map Estimation Network}
\label{sec:depthMapNetwork}
Coded aperture simulated image can be modelled as Eq. \eqref{eqn:simulatedImage}. The Fourier transform of an extracted patch $I^{(p)}$ from the simulated coded aperture image $I$ is the product of the Fourier Transform of blur kernel $C^{(s)}$ with the Fourier transform of the corresponding patch $X^{(p)}$ in the all-in-focus image $X$.
\begin{equation}
    \widetilde I^{(p)} = \widetilde C^{(s)} \widetilde X^{(p)}
\end{equation}



In \cite{levin2007image} it was analyzed that the distinct pattern of zero crossings in the Fourier transform of the defocused image patches reveal the information about the blur kernel size. In order to detect these distinct pattern of zero crossings we use Convolutional Neural Networks (CNN) which are shown to be very efficient in pattern detection and classification tasks. For this particular network we have used 3 convolutional layers followed by 5 fully connected layers. This network takes in as input the Fourier transform of the extracted patches from simulated coded aperture image $I$ and output the blur kernel size $s$. 


\begin{figure}[t]
  \centering
  \centerline{\includegraphics[width=6.5cm, height=6.0cm]{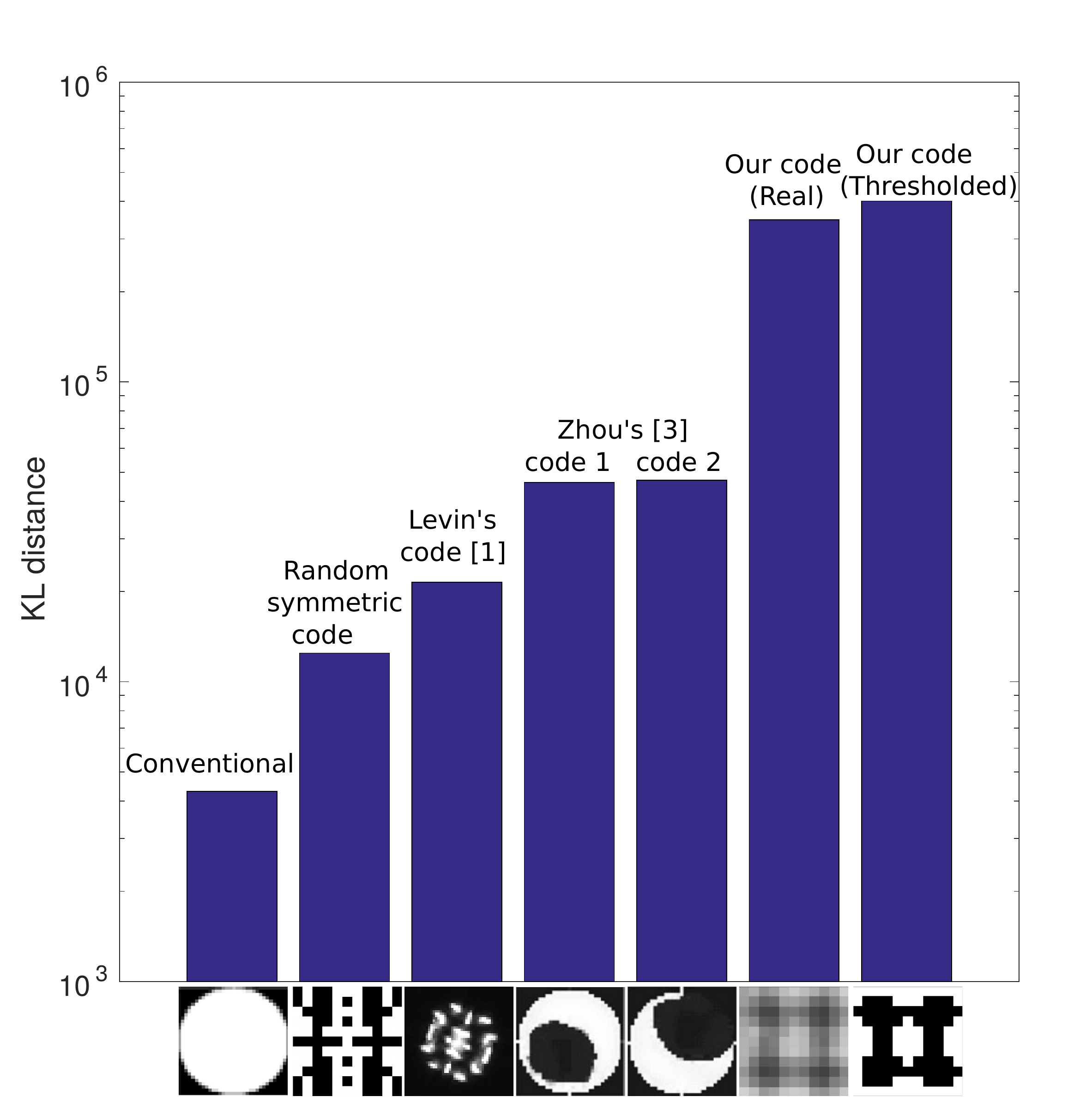}}
\caption{Comparison of KL Distance metric (larger value is better) for different aperture codes. .}
  \label{fig:KL_comparison}
  \vspace{-0.5cm}
\end{figure}
\section{Experiments}
\label{sec:experiments}
We used the NYU Depth Dataset V2 \cite{Silberman:ECCV12} to train our network. The depth maps in the dataset were discretized to blur kernel size according to the following procedure. Thought depth in the real world takes continuous values, the resulting defocus can occupy only a discrete number of pixels on the sensor. The number of pixels occupied by the defocus is the ratio of the blur radius to the pixel pitch. Using the thin lens formula and the depth map, we compute the blur radius for each pixel by assuming practical values for focal length $f$, pixel pitch $\mu$ and the f-stop number $F/\#$. In our experiments, we use $f=25 mm$, $\mu=8\mu m $ and $F/\#=1.4$. Dividing the blur radius by the pixel pitch and rounding off to the nearest integer gives us the size of the blur kernel.

The total dataset was divided into $3$ parts: 60\% for training, $20\%$ for validation and $20\%$ for testing the network. While training patches of size $32 \times 32$ were extracted, blurred with kernel corresponding to the depth in the scene and passed through the depth prediction network. While estimating the depth map, we move with a stride of $8$ with a patch size of $32\times32$ over the whole image. We used the value of $\alpha_t = 2.5 + \frac{t}{3000}$ where t represents the iteration number. In each iteration, a batch of 128 patches is propagated through the network.

\begin{figure*}[t]
  \centering
  \centerline{\includegraphics[width=0.8\textwidth]{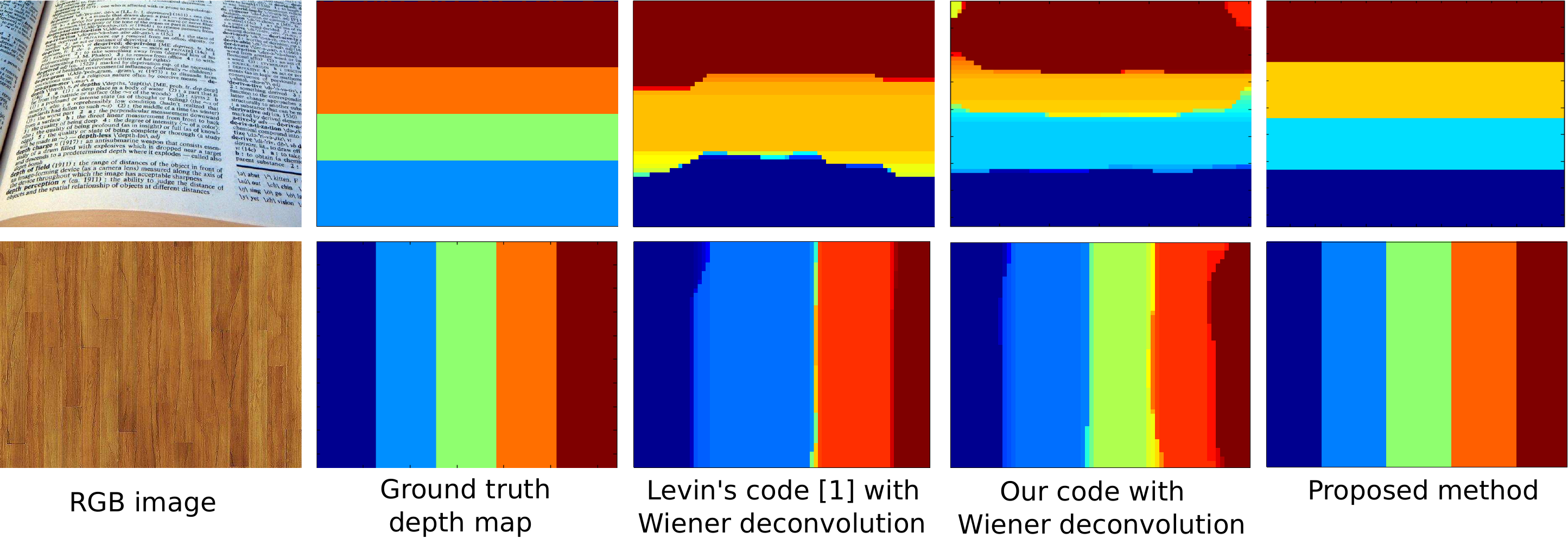}}
  \caption{Code proposed in \cite{levin2007image} fails to give a good depth map, while our code is able to recover accurate depth maps.}
  \label{fig:levin_comparison}
\end{figure*}

\begin{figure*}[ht]
  \centering
  \centerline{\includegraphics[width=0.8\textwidth]{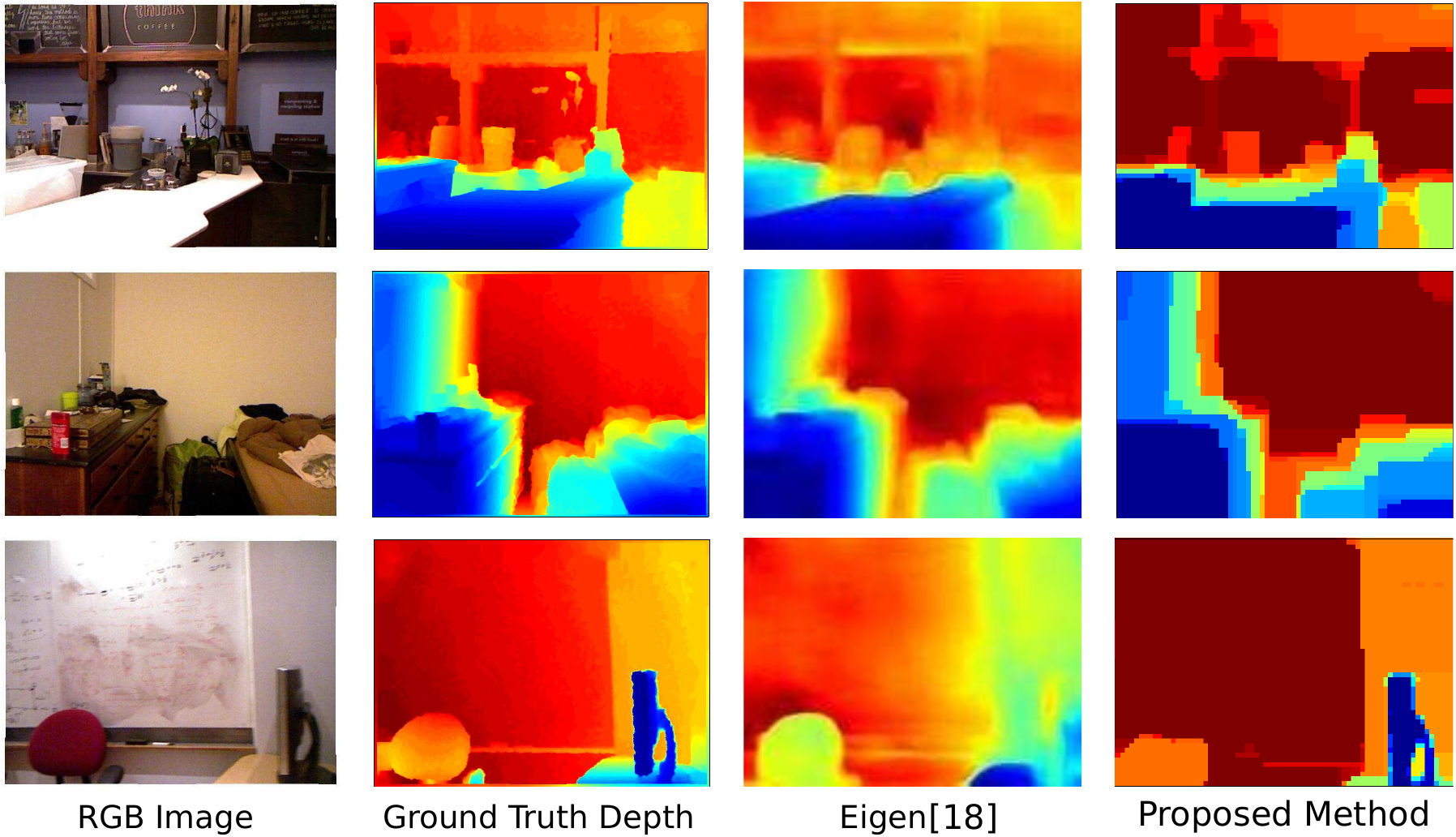}}
  \caption{The figure shows the comparison of the depth map estimated by our network with that of monocular depth map estimation proposed in \cite{eigen2015predicting}. \cite{eigen2015predicting} fails to detect objects as evident from the duster in bottom most row.}
  \label{fig:mnocularcomparison}
  \vspace{-0.5cm}
\end{figure*}
\subsection{Comparison of KL Divergence of Aperture Code}
In \cite{levin2007image}, the authors used KL divergence between the image distributions for images blurred by different blur kernel sizes to quantify the robustness of an aperture filter. The criterion was proposed with the assumption of Gaussian distribution for image gradients of the input image. We evaluate this metric on our code, along with the codes proposed in \cite{levin2007image} and \cite{zhou2009coded}. We also compare a conventional full open aperture and a symmetric random code. Figure \ref{fig:KL_comparison} shows that the our learned code is about $10$ times better than those proposed in \cite{levin2007image,zhou2009coded}. 

\subsection{Depth Map Estimation using Wiener Deconvolution}
Here we demonstrate that our learned code can be used for estimating depth using traditional algorithm similar to as proposed in \cite{levin2007image}. In this algorithm, each patch of the input defocused image is deconvolved (via Wiener deconvolution) with blur kernels of different sizes. These deblurred patches are again convolved with the blur kernels of different sizes and the $l2$ norm is computed with respect to the input defocused patch. The blur kernel size which gives the least $l2$ error is taken as the correct blur size.
We use synthetic images with artificial depth maps for our experiments as shown in fig.\ref{fig:levin_comparison}. We compare our code and Levin's code with traditional depth estimation algorithm as well as our code with our depth prediction network. We can conclude that irrespective of the depth estimation technique, our code performs better than Levin's code.  
\vspace{-0.5cm}
\subsection{Comparison with Monocular Depth Prediction}
We compare our depth map with that of the depth maps obtained from the monocular depth prediction network \cite{eigen2015predicting}, which does not use defocus cues. From Figure \ref{fig:mnocularcomparison}, it is clearly evident that the boundaries are much sharper in our depth map as compared to the depth map predicted by \cite{eigen2015predicting}.  

\section{Conclusion}
\vspace{-0.4cm}
In departure from the previous approaches for coded aperture design, we proposed a data driven approach without explicitly using a predefined prior. We jointly trained the network to evolve the optimal code along with learning the mapping from coded image to depth map. Interestingly, though we haven't optimized the network for any specific metrics, we outperform the KL divergence based performance metric proposed in \cite{levin2007image}. 


\end{document}